\documentclass[runningheads]{llncs}

\usepackage{eccv}

\usepackage{eccvabbrv}

\usepackage{graphicx}
\usepackage{booktabs}
\usepackage[table]{xcolor}
\usepackage{siunitx}

\usepackage{rotating}
\usepackage{subfiles}

\usepackage[accsupp]{axessibility}  

\usepackage{hyperref}

\usepackage{orcidlink}

\newcommand{\perfimprovement}{$4\% - 15\%$}

\begin{document}

\title{From ex(p) to poly: Gaussian Splatting with Polynomial Kernels}

\titlerunning{From ex(p) to poly}

\author{Joerg H. Mueller\inst{1}\orcidlink{0000-0002-6368-6340} \and
Martin Winter\inst{2}\orcidlink{0000-0002-7778-9770} \and
Markus Steinberger\inst{2,3}\orcidlink{0000-0001-5977-8536}}

\authorrunning{J.~H.~Mueller et al.}

\institute{Huawei Technologies, Switzerland \and
Huawei Technologies, Austria \and
Graz University of Technology, Austria \\
\email{\{firstname.lastname\}@huawei.com}}

\maketitle

\begin{abstract}
	Recent advancements in Gaussian Splatting (3DGS) have introduced various modifications to the original kernel, resulting in significant performance improvements.
	However, many of these kernel changes are incompatible with existing datasets optimized for the original Gaussian kernel, presenting a challenge for widespread adoption.
	In this work, we address this challenge by proposing an alternative kernel that maintains compatibility with existing datasets while improving computational efficiency.
	Specifically, we replace the original exponential kernel with a polynomial approximation combined with a ReLU function.
	This modification allows for more aggressive culling of splats, leading to enhanced performance across different 3DGS implementations.
	Our results show a notable performance improvement of \perfimprovement{} with negligible impact on image quality.
	We also provide a detailed mathematical analysis of the new kernel and discuss its potential benefits for 3DGS implementations on NPU hardware.
\end{abstract}

\section{Introduction}
\label{sec:intro}

Recent advances in neural rendering have been significantly accelerated by the introduction of 3D Gaussian Splatting (3DGS)\cite{3DGS2023}, which represents scenes as a collection of anisotropic Gaussian primitives.
In this formulation, each primitive can be interpreted as a combination of a quadric (ellipsoidal support) and an associated kernel function that defines its spatial contribution.
In the original 3DGS formulation, this kernel takes the form of a truncated exponential, corresponding to a Gaussian distribution with finite support for rendering efficiency.

While the Gaussian kernel has proven effective and remains the de facto standard in most 3DGS-based pipelines, it is not the only possible choice.
Several recent works have explored alternative kernel functions to improve numerical stability, rendering quality, or computational efficiency\cite{Huang2DGS2024,chen2024gaussiansfasthighfidelity3d,DBS2025}.
These efforts demonstrate that the kernel component of 3DGS is not inherently fixed, but rather a design choice that can be adapted to specific objectives.

In this work, we revisit the kernel design in 3D Gaussian Splatting with a pragmatic goal: to replace the original Gaussian kernel with a simpler and more computationally efficient approximation.
Evaluating the exponential function represents a significant fraction of the already expensive blending step, as it is a costly operation even using low-precision special functions.
What sets our approach apart from previous work is our focus on maintaining compatibility with existing 3DGS datasets and pipelines.
We specifically target improvements in runtime performance with only minimal degradation in rendering quality on already optimized datasets.
By decoupling the geometric representation from the specific choice of kernel, we demonstrate that a more efficient formulation is possible without sacrificing the advantages that have made 3DGS widely adopted.
Our main contributions are as follows:
\begin{itemize}
	\item We introduce an $N$-th-order polynomial kernel approximation for Gaussian Splatting as a high-performance replacement that is compatible with existing datasets and pipelines, significantly reducing the computational overhead of kernel evaluation without resorting to more expensive Special Function Units.
	We show that a simple first-order polynomial approximation offers the best tradeoff, as it can achieve significant performance gains with negligible quality degradation, while higher-order approximations can further improve the fit at the cost of increased computational complexity if needed.
	\item By leveraging the finite support of the polynomial approximation, we derive a tight, universal bounding radius as well as an even tighter opacity-dependent bounding radius for the first-, second- and third-order polynomials that allow for more aggressive culling of splats, leading to significant performance improvements across various 3DGS implementations.
	\item We provide a formal mathematical derivation proving that anti-aliasing normalization factors remain invariant for arbitrary kernel functions.
	\item We present a methodology for fitting polynomial coefficients using an $L_1$ loss with a sampling strategy that reflects the practical distribution of quadric values in 3DGS, ensuring that the approximation is optimized for real-world rendering scenarios.
	\item We evaluate the new kernel and tighter culling to determine how quality is impacted and how performance is improved in multiple 3DGS implementations across multiple rendering APIs.
\end{itemize}

\section{Preliminaries and Related Work}
We base our work on Gaussian Splatting as introduced by Kerbl et al.\cite{3DGS2023}, which represents a 3D scene as a set of 3D Gaussian splats $G = \{G_1, G_2, ..., G_p\}$.
Each Gaussian splat is represented by its mean $\mu_i \in \mathbb{R}^3$, covariance matrix $\mathbf{\Sigma}_i \in \mathbb{R}^{3 \times 3}$, color $c_i \in \mathbb{R}^3$ and opacity $o_i \in [0, 1]$. 
The rendering is performed by projecting the splats onto the image plane, computing their contribution to each pixel based on their distance to the pixel center $v$ and their opacity
and compositing the contributions in a front-to-back order using alpha blending.
The contribution is computed as
\begin{equation}
		G_i(v) = o_i k\left( Q_i(v) \right) = \exp\left(-\frac{1}{2} \left( v - \mu'_i \right)^T \mathbf{\Sigma}_i ^{\prime-1} \left( v - \mu'_i \right)\right),
\end{equation}%
where $k(Q)$ is a kernel function applied to the quadric $Q_i$ which represents an ellipse based on $\mathbf{\Sigma}'_i$ and $\mu'_i$ after projecting the splat to the image plane.
For a complete description of Gaussian Splatting, please refer to the original paper on 3DGS\cite{3DGS2023}.
In addition, we also opted to evaluate our polynomial kernels with the tile culling as introduced by StopThePop\cite{radl2024stopthepop}
and also investigate anti-aliasing techniques as introduced by Mip-Splatting \cite{Yu2024MipSplatting} and AAA-Gaussians\cite{steiner2025aaags}.

\subsection{Tight Tile Culling}
One crucial step in optimizing performance for Gaussian splatting is tight culling of tiles based on the opacity of the splats to reduce overdraw. 
The original formulation determines the affected tiles per splat $G_i$ based on the square inscribing the circle defined by the center $\mu_i$ and the radius $r = 3 \sqrt{\lambda_{max}}$ (based on the largest eigenvalue of the projected covariance matrix).
This not only neglects the opacity of the splat but generally overestimates the affected area significantly, especially for highly anisotropic splats.
Several works have proposed methods for tighter culling, GCCS \cite{10.1145/3725843.3756072} improves upon the original method by calculating an exact bound based on the splat's opacity and GSCore \cite{10.1145/3620666.3651385} further generates a tighter, rectangular bound.
StopThePop\cite{radl2024stopthepop}, Speedy-Splat\cite{HansonSpeedy} and FlashGS\cite{11093809} generate even tighter bounds by following a two stage design, 
generating a tighter rectangular bound first and then evaluating the point of maximum contribution within each tile to determine whether the tile is affected or not.
We opt to evaluate the tighter tile culling as presented in StopThePop\cite{radl2024stopthepop} due to its efficiency.

\subsection{Alternative Splatting Kernels}
Since the introduction of Gaussian Splatting\cite{3DGS2023}, several works have proposed alternative splatting kernels to improve surface fitting, reduce floaters and better represent sharp edges.
Recent literature has focused on replacing the Gaussian kernel with more flexible or geometrically accurate primitives to improve reconstruction quality and efficiency.
It is important to note though that most of these approaches are incompatible with existing datasets optimized for the original Gaussian kernel, which is a significant barrier to their adoption in practice.
\subsubsection{Geometric and Surface Alignment}
To better represent surfaces, 2D Gaussian Splatting~\cite{Huang2DGS2024} flattens the Gaussians along the viewing direction to create 2D ellipses that better align with surfaces.
TNT-GS~\cite{10.1145/3746027.3755573} expands on this by truncating 2D Gaussians (with a learnable truncation radius) and stacking multiple truncated Gaussians along the viewing direction to form a more accurate representation of surfaces while reducing floaters.
Similarly, 3D Convex Splatting~\cite{held20253dconvexsplattingradiance} replaces Gaussians with convex polytopes to better approximate scene geometry, while Quadratic Gaussian Splatting~\cite{zhang2025quadraticgaussiansplattinghigh} uses quadratic functions for splatting to improve surface fitting and reduce floaters.
Triangle Splatting~\cite{Held2025Triangle} avoids volumetric integration entirely by optimizing explicit triangles, allowing for seamless integration with traditional mesh-based rendering techniques.

\subsubsection{Sharp Edges and Frequencies}
Standard Gaussians often struggle with sharp discontinuities. 
3D Half-Gaussian Splatting~\cite{3DHGS2025} introduces a half-Gaussian kernel, where each half can have different opacity values, to better model sharp edges, while Deformable Radial Kernel Splatting~\cite{DRKS2025} employs radial basis functions for better edge representation while also capturing more complex structures with fewer primitives.
3D Linear Splatting~\cite{chen2024gaussiansfasthighfidelity3d} uses a linear "tent" kernel to capture sharp features more effectively, also demonstrating that shorter-tailed kernels can reduce blurring artifacts in high-frequency regions.
Gabor Splatting ~\cite{10.1145/3641234.3671081} utilizes Gabor functions to represent both spatial and frequency information, enhancing texture representation in splatting.
A different approach is employed by Disc-GS~\cite{10.5555/3737916.3741482}, which retains the Gaussian distribution, but uses cubic Bèzier curves to cut the projected Gaussians to better represent sharp discontinuities.

\subsubsection{Generalized Distributions}
Several methods generalize the mathematical formulation of the splat. Generalized Exponential Splatting~\cite{GES2024} introduces a shape parameter to control the falloff sharpness of the primitive, allowing it to adapt between Gaussian and rectangular shapes.
3D Student Splatting~\cite{zhu20253d} employs the Student's t-distribution for splatting, which has heavier tails than the Gaussian, enabling better representation of fine details and reducing floaters.
Deformable Beta Splatting~\cite{DBS2025} uses a Beta kernel that can morph between box-like and Gaussian-like shapes by adjusting its shape parameters, 
while Universal Beta Splatting~\cite{liu2025universalbetasplatting} extends this concept to n-dimensional Beta kernels, allowing simultaneous modeling of position, viewing direction, time, and other properties within a unified mathematical framework.
DARB-Splatting~\cite{arunan2025darbsplattinggeneralizingsplattingdecaying} further generalizes this concept by introducing Decaying Anisotropic Radial Basis Functions, allowing for a broader class of reconstruction kernels.
Don't Splat your Gaussians~\cite{10.1145/3711853} proposes using the Epanechnikov kernel (finite support) as an alternative, demonstrating improved performance in volumetric ray-tracing frameworks.

\section{Method}

In this paper, we approximate the exponential kernel function $g(x) = \exp\left(-\frac{1}{2}x\right)$ used on the quadric $x = Q_i\left(v\right)$ in Gaussian Splatting with a ReLU-activated $N$-th-order polynomial
\begin{equation}
	f_N(x) = \max\left(\sum_{i=0}^N c_i x^i; 0\right).
\end{equation}%
In the simplest case where we expect the most performance gain, the approximation uses a simple first-order polynomial, \ie, $f_1(x) = \max\left(c_1x+c_0; 0\right)$.
In this section we will discuss the optimization process for the coefficients $c_i$ and how this modified kernel affects culling and anti-aliasing.

\subsection{Approximation of the Exponential Function}

The exponential function has infinite support, but, for practical rendering purposes, 3DGS limits the support by cutting the function off at $g(x) < \epsilon$, with the threshold $\epsilon = \frac{1}{255}$~\cite{3DGS2023}.
As the quadric $Q_i(v) >= 0$, the relevant support of our approximation is in the range $0 \le x \le -2 \ln(\epsilon)$.
We optimize the parameters $c_i$ in the relevant value range for $x$ using an $L_1$ loss and gradient descent.

A crucial factor for the optimization is the sample distribution which should mirror the sample distribution that we have in practice.
Since the splats in 3DGS are sampled in 2D, the sampling should reflect the pixel-wise 2D sampling in screen space.
The linear transformation of the covariance matrix allows the quadric to be mapped to the unit disk keeping uniform area sampling uniform.
The unit disk can be sampled uniformly in area by sampling the polar coordinates uniformly in the angle $\theta \sim \text{U}(0,2\pi)$ and uniformly in the squared radius $\rho^2 \sim \text{U}(0,1)$.
Consequently, this leads to a uniformly sampled quadric $x \sim \text{U}(0,-2 \ln(\epsilon))$ in the area of interest.
This is also true if the splats are evaluated by their maximum contribution in 3D as the sampled primitive is a two-dimensional surface.
In contrast, would the splat be sampled volumetrically, the sampling density would need to be proportional to the square $\propto x^2$.

Fitting the first-order polynomial approximation $f_1(x)$ to the function $g(x)$ using uniform sampling in the range $x \sim \left[0, -2\ln\left(\frac{1}{255}\right)\right]$ leads to parameters $c_0 \approx -0.176$ and $c_1 \approx 0.773$.
Notably, rendering with this, the maximum opacity does not reach one and $f_1(x) = 0$  for $x > 4.386$, which corresponds to culling the exponential at a value of $t = \frac{28.45}{255}$ or at $2.1 \sigma$, much tighter than 3DGS at $3.3 \sigma$.
Before looking at higher-order fits, we discuss culling for $f_1(x)$.

\subsection{Culling}

Since the kernel we fit reaches zero faster, we can cull the splats more tightly at the specific bound without introducing artifacts, significantly reducing the number of tiles that need to be evaluated.
Radl et al.\cite{radl2024stopthepop} found that culling can be made tighter based on the opacity and the cut-off threshold $\epsilon$ to

\begin{equation}
	t_{g} = \sqrt{2 \ln\left(\frac{o}{\epsilon}\right)}.
\end{equation}%
For our first-order polynomial approximation we can compute this bounding as

\begin{equation}
	t_{f_1} = \sqrt{\frac{\epsilon - oc_0}{oc_1}},
\end{equation}%
which, if we set $\epsilon = 0$ for the zero crossing, is simplified to
\begin{equation}
	\label{eq:tight_bound_linear}
	t'_{f_1} = \sqrt{-\frac{c_0}{c_1}},
\end{equation}%
which is independent of the opacity $o$, resulting in a universal $t'_{f_1}$ for all splats that can be precomputed ahead of time.

\subsection{Higher-Order Polynomials}
\begin{figure}
	\centering
	\includegraphics[width=\textwidth]{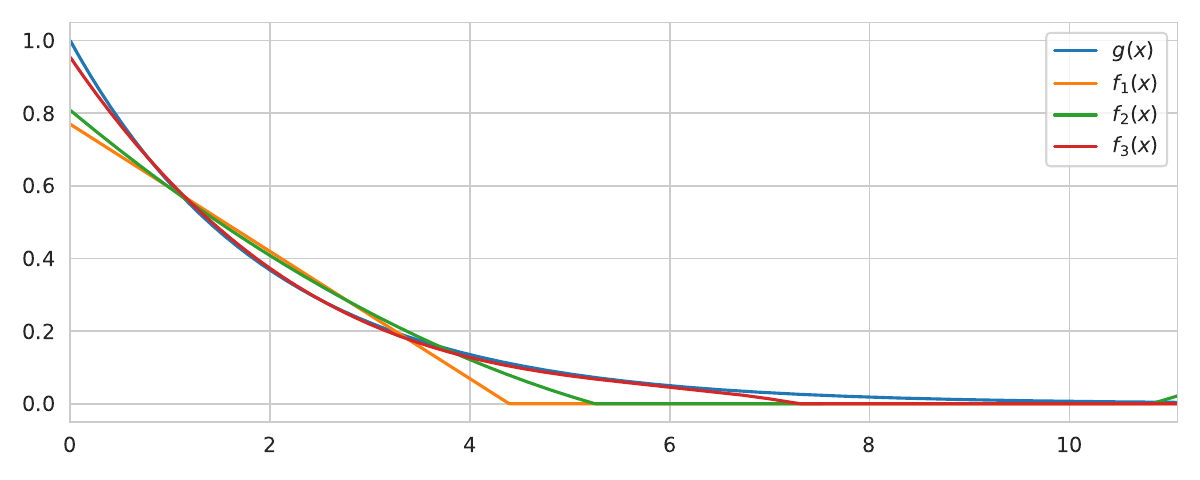}
	\caption{Different kernel functions for Gaussian Splatting depending on the quadric $x$.
	$g(x)$ is the usual exponential function and $f_N(x)$ are polynomial approximations of order $N$ approximated with gradient descent on the $L_1$ loss in comparison to $g(x)$.
	The plot shows exactly the relevant range of the kernel function for $x \sim \left[0, -2\ln\left(\frac{1}{255}\right)\right]$.}
	\label{fig:fits}
\end{figure}

We further extend our investigation to higher-order approximations, specifically quadratic and cubic polynomials. 
While the increased instruction count of these kernels may limit their practical utility compared to the first-order case, they still offer valuable insights. 
The quadratic fit in Figure~\ref{fig:fits} shows a significant challenge with higher-order fits, which is the lack of monotonicity; because the parabola eventually becomes positive again outside the intended fitting range. 
To maintain the expected decay of the Gaussian kernel, we propose two strategies to mitigate this behavior.
One option is to modify the kernel function to
\begin{equation}
	\label{eq:quadratic_if}
	f'_N(x) = \begin{cases}
		\sum_{i=0}^N c_i x^i & \text{if } x < x_0 \\
		0 & \text{otherwise}
	\end{cases},
\end{equation}%
where $x_0$ is the first root of the polynomial where it becomes zero.
The other option is to adjust the fitting range to the maximum quadric value being evaluated, which depends on which kind of culling is used.
A derivation of this approach for a tile-based culling can be found in the supplementary material.

When employing higher-order approximations, careful consideration of the polynomial's critical points is essential. 
The presence of local extrema within the target range can introduce visual artifacts, as the opacity doesn't decrease monotonically with radial distance as it does for the exponential function.
Fitting a third-order polynomial over the interval of interest results in a strictly monotonic cubic function with a single real root. 
Consequently, the third-order kernel requires no additional modifications to ensure stability. 
Furthermore, the roots of both the second- and third-order polynomials provide well-defined boundaries for tighter culling; the corresponding analytical derivations are provided in the supplementary material.

\subsection{Anti-Aliasing}
Yu et al.\cite{Yu2024MipSplatting} demonstrated that consistent anti-aliasing requires more than covariance regularization, i.e. adding a constant to the diagonal, as proposed in the original 3DGS framework \cite{3DGS2023}.
Instead, a normalization factor is essential to ensure that the contribution of the splat remains constant across different views.
In 3DGS, anti-aliasing is modeled by convolving the unnormalized Gaussian PDF with a Gaussian low-pass filter.
This operation, representing the convolution of two normal distributions, results in another normal distribution with the means and covariance matrices added together.
Because the standard 3DGS kernel lacks the $\sqrt{\left(2\pi\right)^N \left|\Sigma\right|}$ normalization factor of the Gaussian distribution, the normalization for correct filtering cancels the $\sqrt{\left(2\pi\right)^N}$ term, leaving $\sqrt{\left|\Sigma\right|}$.

While recent works have applied this anti-aliasing logic to alternative kernels, they often do so without a formal discussion of its validity.
Given that the quadric inside the kernel represents an ellipsoid, adding a value $v$ to the diagonal of the covariance matrix effectively scales the ellipsoid's axes by $\sqrt{s_i^2 + v^2}$\cite{steiner2025aaags} for each axis $i$, independent of the chosen kernel function $k$.
While this axis scaling is inherently kernel-agnostic, the question remains whether the same normalization factor can be applied across different kernels.
To compute the normalization factor $N_\Sigma$, we compute the integral over the kernel function as
\begin{equation}
	N_\Sigma = \int_{\mathbb{R}^N} k\left(\left(x - \mu\right)^T \Sigma^{-1} \left(x - \mu\right)\right) dx,
\end{equation}%
where we substitute $x = M^T y + \mu$, where $\Sigma = M^TM$ is the decomposition of the positive semi-definite covariance matrix.
Using the differential
\begin{equation}
	d^Nx = \left|M^T\right| d^Ny = \sqrt{\left|\Sigma\right|} d^Ny,
\end{equation}%
we arrive at
\begin{align}
	N_\Sigma &= \int_{\mathbb{R}^N} k\left(\left(M^T y + \mu - \mu\right)^T \left(M^TM\right)^{-1} \left(M^T y + \mu - \mu\right)\right) \sqrt{\left|\Sigma\right|} dy \nonumber \\ 
	&= \sqrt{\left|\Sigma\right|} \int_{\mathbb{R}^N} k\left(y^Ty\right) dy,
\end{align}%
where we can see that indeed, we get the same normalization factor dependence on $\Sigma$ and therefore can indeed apply the same normalization for any kernel function $k$ and thus use the ratio of the normalization factors of the unaltered and altered covariance matrices.
The remaining factor for the normalization is independent of the quadric and only depends on the kernel itself, thus canceling in a ratio of two normalization factors as used for anti-aliasing.
Therefore, using the same technique as MIP-splatting\cite{Yu2024MipSplatting} can be used as an anti-aliasing method effectively, by adapting the size of the ellipsoid or ellipse ensuring a minimum scale, but unlike for the Gaussian kernel, this method is not a convolution-based filter as in general $k_a(Q(\mu, \Sigma + v I)) \neq k_a(Q(\mu, \Sigma)) \ast k_b(Q(0, v I))$ for two normalized kernels $k_a$ and $k_b$, unless both of them are Gaussian.

\section{Evaluation}
Following the evaluation protocol of Kerbl et al.\cite{3DGS2023}, we evaluate the quality and performance of our method on the Mip-NeRF 360~\cite{barron2022mipnerf360}, Tanks and Temples~\cite{Knapitsch2017} and Deep Blending~\cite{DeepBlending2018} datasets.
Performance was benchmarked on a workstation equipped with an RTX 5090 and on a Mac featuring an M1 Ultra for MetalSplatter\cite{cier2024metalsplatter}.
In all experiments, anti-aliasing was enabled to maintain consistency and ensure high-fidelity results.
Unless otherwise mentioned, we evaluate with a slightly modified version of vanilla 3DGS\cite{3DGS2023} where we added the tile culling from StopThePop\cite{radl2024stopthepop} as our Baseline, which was also used for the optimization of all scenes/models.

\subsection{Quality}
\label{sec:quality}

To evaluate the quality of the different kernels, we extended Baseline with the different optimized kernels of orders 1-3 and Mip-Splatting-like 2D anti-aliasing.
Following the optimization protocol established by Kerbl et al. \cite{3DGS2023}, scenes were trained for \num{30000} iterations using the exponential kernel $g(x)$.
Quality was assessed on the test set using PSNR, SSIM, and L-PIPS metrics, with results summarized in Table~\ref{tab:quality}.

\begin{table}[b!]
	\centering
	\caption{We compare the quality of the different kernels using the quality metrics PSNR, SSIM and LPIPS.
	For LPIPS lower is better, for the other two higher is better.
	The higher the order of the polynomial, the better the quality approaches that of $g(x)$, or even surpases it for $f_3(x)$.}
	\label{tab:quality}
	\begin{tabular}{l|rrrr|rrrr|rrrr}
\toprule
 & \multicolumn{4}{c}{PSNR $\uparrow$} & \multicolumn{4}{c}{SSIM $\uparrow$} & \multicolumn{4}{c}{LPIPS $\downarrow$} \\
 & $g(x)$ & $f_1(x)$ & $f'_2(x)$ & $f_3(x)$ & $g(x)$ & $f_1(x)$ & $f'_2(x)$ & $f_3(x)$ & $g(x)$ & $f_1(x)$ & $f'_2(x)$ & $f_3(x)$ \\
\midrule
bicycle & 25.06 & 24.47 & 24.77 & 25.05 & 0.749 & 0.734 & 0.742 & 0.749 & 0.243 & 0.249 & 0.244 & 0.242 \\
bonsai & 32.43 & 30.89 & 31.58 & 32.41 & 0.947 & 0.932 & 0.940 & 0.947 & 0.180 & 0.191 & 0.183 & 0.178 \\
counter & 29.10 & 28.28 & 28.65 & 29.07 & 0.916 & 0.900 & 0.907 & 0.915 & 0.182 & 0.191 & 0.186 & 0.182 \\
drjohnson & 29.43 & 29.12 & 29.31 & 29.44 & 0.906 & 0.898 & 0.902 & 0.906 & 0.235 & 0.238 & 0.235 & 0.234 \\
flowers & 21.41 & 20.97 & 21.19 & 21.38 & 0.586 & 0.574 & 0.581 & 0.586 & 0.362 & 0.360 & 0.358 & 0.360 \\
garden & 27.29 & 26.34 & 26.79 & 27.26 & 0.854 & 0.835 & 0.845 & 0.854 & 0.130 & 0.137 & 0.133 & 0.129 \\
kitchen & 31.62 & 29.71 & 30.58 & 31.57 & 0.933 & 0.912 & 0.922 & 0.933 & 0.115 & 0.132 & 0.124 & 0.115 \\
playroom & 30.21 & 29.95 & 30.13 & 30.24 & 0.910 & 0.903 & 0.906 & 0.910 & 0.239 & 0.242 & 0.239 & 0.238 \\
room & 31.77 & 30.13 & 30.74 & 31.67 & 0.928 & 0.913 & 0.920 & 0.927 & 0.196 & 0.205 & 0.200 & 0.195 \\
stump & 26.56 & 25.92 & 26.24 & 26.54 & 0.767 & 0.750 & 0.760 & 0.767 & 0.244 & 0.250 & 0.245 & 0.243 \\
train & 22.08 & 21.39 & 21.69 & 22.04 & 0.819 & 0.803 & 0.812 & 0.819 & 0.202 & 0.206 & 0.202 & 0.201 \\
treehill & 22.69 & 22.36 & 22.53 & 22.67 & 0.637 & 0.624 & 0.631 & 0.637 & 0.349 & 0.356 & 0.351 & 0.348 \\
truck & 25.45 & 24.41 & 24.85 & 25.40 & 0.884 & 0.865 & 0.875 & 0.884 & 0.148 & 0.153 & 0.150 & 0.147 \\
\bottomrule
\end{tabular}

\end{table}

\begin{figure}
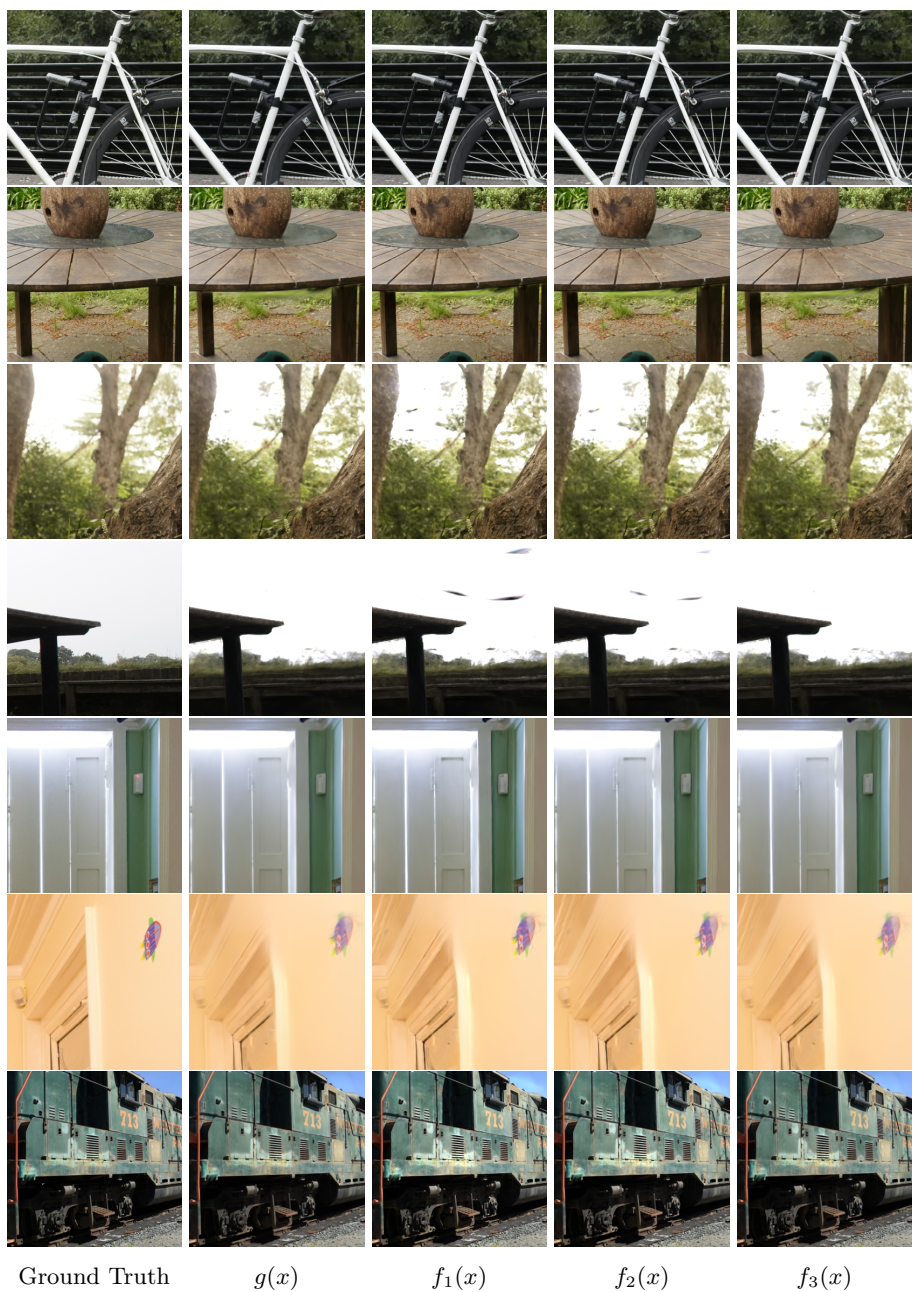

	\centering
	\include{samples}
	\vspace{-2em}
	\begin{minipage}[t]{0.19\textwidth}
		\centering
		Ground Truth
	\end{minipage}
	\begin{minipage}[t]{0.19\textwidth}
		\centering
		$g(x)$
	\end{minipage}
	\begin{minipage}[t]{0.19\textwidth}
		\centering
		$f_1(x)$
	\end{minipage}
	\begin{minipage}[t]{0.19\textwidth}
		\centering
		$f_2(x)$
	\end{minipage}
	\begin{minipage}[t]{0.19\textwidth}
		\centering
		$f_3(x)$
	\end{minipage}
	\caption{Sample render cut-outs of the test dataset taken from some of the test scenes, rendered using the different kernels.
	In most cases, the different kernels are indistinguishable.
	Only scenes with white backgrounds (3rd and 4th row) show noticeable artifacts for $f_1(x)$ and $f_2(x)$ that we discuss in section~\ref{sec:limitations}.}
	\label{fig:samples}
\end{figure}

Rendering quality scales with the polynomial order's ability to fit the exponential function.
Notably, the third-order polynomial produces results comparable to the exponential Baseline, and in some instances, outperforms it, suggesting that higher-order expansion yield diminishing returns.
As illustrated in Figure~\ref{fig:samples}, any reduction in quality across the tested orders is practically negligible.

\subsection{Performance}
To demonstrate the universal applicability of our method, we integrate it into five implementations of Gaussian Splatting: Baseline \cite{3DGS2023,radl2024stopthepop}, gsplat \cite{ye2025gsplat}, Faster-GS\cite{hahlbohm2026fastergs}, vk\_gaussian\_splatting \cite{vk_gaussian_splatting} and MetalSplatter \cite{cier2024metalsplatter}.
Baseline is a modified vanilla 3DGS implementation with added tile culling from StopThePop~\cite{radl2024stopthepop}, which is also used by Faster-GS\cite{hahlbohm2026fastergs}.
Table~\ref{tab:timing} shows the performance of each approach in comparison to our optimized version, utilizing a first-order polynomial approximation and opacity-dependent culling.

\begin{table}[b!]
	\centering
	\caption{Performance measurements of our first-order approximation in different Gaussian Splatting implementations.
	Timing measurements in the kernel columns ($g(x)$ and $f_1(x)$) are the frame time in milliseconds, measured on the GPU, except for MetalSplatter~\cite{cier2024metalsplatter} where sorting is done on the CPU and we therefore measured CPU frame time.
	The \% column shows the performance improvement of $f_1(x)$ over $g(x)$ in percent.}
	\label{tab:timing}
	\begin{tabular}{l|rrr|rrr|rrr|rrr|rrr}
\toprule
 & \multicolumn{3}{c}{Baseline} & \multicolumn{3}{c}{gsplat} & \multicolumn{3}{c}{Faster-GS} & \multicolumn{3}{c}{Vulkan} & \multicolumn{3}{c}{Metal} \\
 & $g(x)$ & $f_1(x)$ & \hspace{0.5em}\% & $g(x)$ & $f_1(x)$ & \hspace{0.5em}\% & $g(x)$ & $f_1(x)$ & \hspace{0.5em}\% & $g(x)$ & $f_1(x)$ & \hspace{0.5em}\% & $g(x)$ & $f_1(x)$ & \hspace{0.5em}\% \\
\midrule
bicycle & 3.74 & 3.29 & 12 & 3.05 & 2.74 & 10 & 2.55 & 2.38 & 7 & 2.49 & 2.21 & 11 & 16.60 & 14.35 & 14 \\
bonsai & 2.40 & 2.07 & 14 & 1.19 & 0.93 & 21 & 0.85 & 0.75 & 11 & 0.79 & 0.62 & 22 & 5.11 & 3.89 & 24 \\
counter & 2.24 & 1.59 & 29 & 1.65 & 1.28 & 23 & 1.01 & 0.87 & 14 & 1.12 & 0.87 & 22 & 6.63 & 4.74 & 29 \\
drjohnson & 2.43 & 2.03 & 17 & 2.03 & 1.71 & 16 & 1.20 & 1.10 & 8 & 1.50 & 1.28 & 15 & 10.18 & 7.87 & 23 \\
flowers & 3.00 & 2.38 & 21 & 1.75 & 1.50 & 14 & 1.57 & 1.43 & 9 & 1.49 & 1.34 & 10 & 10.28 & 8.88 & 14 \\
garden & 2.67 & 2.15 & 20 & 2.63 & 2.31 & 12 & 2.37 & 2.21 & 7 & 2.33 & 2.14 & 8 & 15.60 & 13.68 & 12 \\
kitchen & 2.39 & 1.96 & 18 & 2.21 & 1.77 & 20 & 1.64 & 1.46 & 11 & 1.27 & 1.00 & 21 & 7.38 & 5.75 & 22 \\
playroom & 1.76 & 1.42 & 19 & 1.61 & 1.23 & 24 & 0.87 & 0.76 & 13 & 1.19 & 0.93 & 21 & 6.86 & 5.58 & 19 \\
room & 2.61 & 1.93 & 26 & 1.53 & 1.18 & 23 & 0.92 & 0.78 & 15 & 1.44 & 0.90 & 38 & 5.87 & 4.23 & 28 \\
stump & 2.95 & 2.54 & 14 & 1.93 & 1.68 & 13 & 1.43 & 1.32 & 8 & 1.39 & 1.26 & 10 & 13.93 & 13.15 & 6 \\
train & 2.06 & 1.78 & 14 & 1.72 & 1.41 & 18 & 1.21 & 1.12 & 7 & 1.48 & 1.11 & 25 & 8.24 & 5.69 & 31 \\
treehill & 2.62 & 2.21 & 15 & 1.84 & 1.55 & 16 & 1.52 & 1.36 & 10 & 1.46 & 1.31 & 10 & 11.90 & 10.49 & 12 \\
truck & 1.87 & 1.64 & 12 & 2.07 & 1.77 & 15 & 1.42 & 1.36 & 4 & 1.73 & 1.47 & 15 & 10.53 & 8.19 & 22 \\
average & 2.52 & 2.08 & 18 & 1.94 & 1.62 & 17 & 1.43 & 1.30 & 10 & 1.51 & 1.26 & 18 & 9.93 & 8.19 & 20 \\
\bottomrule
\end{tabular}

\end{table}

Our approach yields substantial performance improvements across all tested implementations.
Among the CUDA-based frameworks - Baseline, gsplat and Faster-GS - which utilize screen-space tile partitioning, the most pronounced gains are observed in Baseline and gsplat.
Notably, even in Faster-GS (the most optimized version of the three), our approximation still achieves performance gains of 4\% to 15\%.
The remaining two frameworks, based on Vulkan and Metal, rasterize an oriented bounding rectangle encompassing the ellipse, which can be reduced in size based on our tighter culling methods.
The performance gains for these rasterization-based methods indicate that our improvements are effective regardless of the underlying rendering architecture.

\subsection{Ablation}

Table~\ref{tab:ablation} shows an ablation study isolating the effects of our polynomial approximation and the two proposed culling variations.
Results reflect average performance across all scenes for Baseline, with additional frame time measurements for Faster-GS.
While replacing the exponential function with a ReLU-polynomial combination already yields a slight performance gain for Baseline, 
it results in a performance regression for the already optimized Faster-GS, which already utilizes hardware-accelerated approximations of the exponential.
The primary driver of performance improvement is our tighter culling strategy.
Specifically, culling based on the polynomial's zero crossing significantly improves performance, which itself is again bested by opacity-aware culling using the same $\frac{1}{255}$ threshold as Gaussian Splatting.
As discussed in Section~\ref{sec:quality}, the polynomial approximation leads to a slight decrease in quality across all metrics, which is not affected by the tighter culling, preserving the same visual fidelity.
While higher-order polynomials improve quality over first-order polynomials - and remain faster on average than the exponential kernel when paired with opacity-aware culling - their performance gains are inconsistent across scenes, with some instances showing worse performance than the exponential.
Only the first-order polynomial $f_1(x)$ consistently provides a performance improvement over the exponential function, due to its minimal computational overhead and tightest culling bounds.
Per-scene timing details are available in the supplementary material for further analysis.

Finally, we investigate the impact of optimizing scenes directly with our polynomial approximation, rather than merely using it as a drop-in replacement during inference.
To support this, we modified the backward pass of Baseline to compute the derivatives of the ReLU-polynomial combination.
Direct optimization significantly alters the resulting splat density and size distribution, which in turn affect both quality and performance.
As shown in Table~\ref{tab:ablation}, optimization improves SSIM and PSNR, though LPIPS metrics exhibit a slight decline. 
While optimization further enhances rendering throughput, we prioritize compatibility with existing datasets trained on the standard exponential kernel; consequently, we do not explore this further, leaving it as an avenue for future work. 
Comprehensive per-scene results are provided in the supplementary material.

\begin{table}
	\centering
	\caption{We ablate our methods separating the influence of the changed kernel and culling, computing quality metrics (for SSIM and PSNR higher is better, for LPIPS lower is better) and average frame times for Baseline and Faster-GS.
	Values are averaged over all evaluated scenes.
	Culling (Cul) is done with either unaltered StopThePop\cite{radl2024stopthepop} culling (S), culling to the root of the approximation polynomial (0) or to the tighter value of $\frac{1}{255}$ (1).
	We also evaluate optimization (Opt) of the scenes with the shown kernel directly instead of using a scene optimized with $g(x)$.}
	\label{tab:ablation}
	\begin{tabular}{lllrrrrr}
\toprule
Kernel & Cul & Opt & SSIM $\uparrow$ & PSNR $\uparrow$ & LPIPS $\downarrow$ & Baseline & Faster-GS \\
\midrule
$g(x)$ & S &  & 0.834 & 27.316 & 0.217 & 2.468 & 1.427 \\
$f_1(x)$ & S &  & 0.819 & 26.456 & 0.224 & 2.375 & 1.447 \\
$f_1(x)$ & 0 &  & 0.819 & 26.456 & 0.224 & 2.207 &  \\
$f_1(x)$ & 1 &  & 0.819 & 26.456 & 0.224 & 2.018 & 1.300 \\
$f'_2(x)$ & S &  & 0.827 & 26.851 & 0.219 & 2.390 & 1.484 \\
$f'_2(x)$ & 1 &  & 0.827 & 26.851 & 0.219 & 2.159 & 1.360 \\
$f_3(x)$ & S &  & 0.833 & 27.289 & 0.216 & 2.428 & 1.455 \\
$f_3(x)$ & 1 &  & 0.833 & 27.289 & 0.216 & 2.364 & 1.416 \\
$f_1(x)$ & 1 & \checkmark & 0.826 & 27.125 & 0.227 & 1.972 &  \\
$f'_2(x)$ & 1 & \checkmark & 0.829 & 27.187 & 0.224 & 2.016 &  \\
$f_3(x)$ & 1 & \checkmark & 0.833 & 27.301 & 0.217 & 2.272 &  \\
\bottomrule
\end{tabular}

\end{table}

\section{Discussion}

Our evaluation demonstrates that a first-order polynomial approximation of the exponential function is a practical choice for real-world applications.
It offers a modest yet potentially significant performance improvement with only a negligible reduction in quality, which is hardly noticeable in actual rendered images.

\subsection{Neural Processing Units}

One reason our approach uses a polynomial together with a ReLU is that these methods can be efficiently implemented on a Neural Processing Unit (NPU).
TC-GS\cite{TCGS2025} shows that the evaluation of the quadric can be reformulated as a matrix multiplication across multiple pixels and splats at the same time.
By decomposing the quadric into two 6-dimensional vectors, $u$ (pixel-dependent) and $v$ (splat-dependent), decoupling the expression into the dot product of two individual vectors.
Furthermore, the opacity $o$ and the constant factor $-\frac{1}{2}$ of the splats are integrated into the exponent directly, by taking the logarithm of the opacity and thus putting it inside the exponential function: $o \exp(-\frac{1}{2}x) = \exp(\ln(o) - \frac{1}{2}x)$.
This transformation allows the entire argument of exponential function to be computed as a single inner product.

This approach is equally applicable to polynomial computation, especially since NPU hardware often has a very efficient ReLU functionality that can be applied immediately following matrix multiplication.
Since the opacity is positive, it can be moved inside the ReLU to scale the entire polynomial.
For the first-order polynomial, we can modify the components of the splat vector $v$ of TC-GS as $v_{0}' = o c_1 \left(\mu_x^{\prime 2}\sigma_{11} +2\mu'_x\mu'_y\sigma_{12}+\mu_y^{\prime 2}\sigma_{22}\right) + c_0$ and $v_i' = -2 o c_1 v_i$ for $i = 1 .. 5$ to get our modified splat vector $v'$.
For higher-order polynomials, the dimensionality of the vector increases with $n_c = \binom{2 \left(N + 1\right)}{2}$, where $N$ denotes the order and $n_c$ the number of components.
Note that since this approach does not compute $x$ directly, the modification done in Eq.~\ref{eq:quadratic_if} is not easily possible and therefore extending the fitting range based on culling might be necessary.

\subsection{Limitations}
\label{sec:limitations}

\begin{figure}[t]

	\centering
	\begin{subfigure}{0.49\linewidth}
		\includegraphics[width=\linewidth]{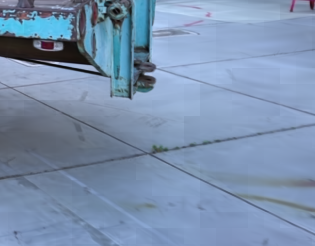}
		\caption{$g(x)$ with opacity-aware $f_1(x)$ culling to $\frac{1}{255}$}
		\label{fig:expartifacts}
	\end{subfigure}
	\hfill
	\begin{subfigure}{0.49\linewidth}
		\includegraphics[width=\linewidth]{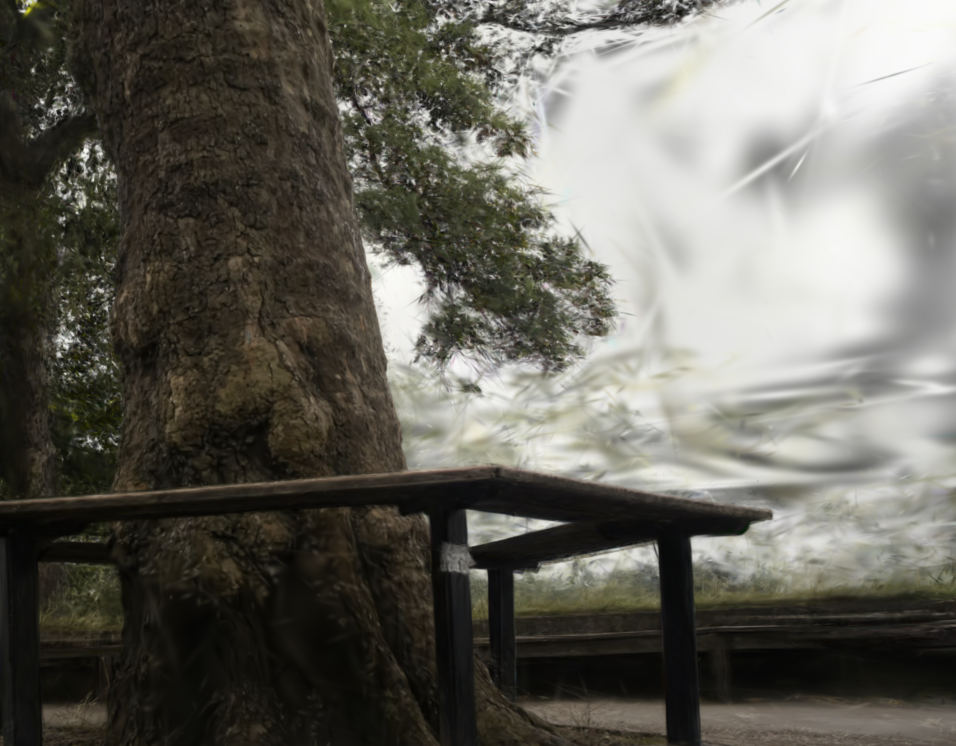}
		\caption{\emph{treehill} rendered by vk\_gaussian\_splatting}
		\label{fig:vktreehill}
	\end{subfigure}
	\caption{Artifacts caused by different mechanisms.
	(a) shows the \emph{truck} scene rendered with $g(x)$ but with the opacity-aware culling of $f_1(x)$, culling to $\frac{1}{255}$.
	In contrast to using $f_1(x)$, using $g(x)$ introduces noticeable blocky artifacts especially for bigger splats as the culling is too tight and the opacity at the culling edge not low enough.
	(b) shows what happens when the \emph{treehill} scene is color clamped before blending as happens in vk\_gaussian\_splatting.
	This also happens in vanilla Gaussian Splatting when an upper limit on the color is introduced.}
\end{figure}

The performance gain realized by substituting the exponential function is inherently constrained.
While approximations typically entail a slight reduction in quality metrics, the perceptual impact is negligible as the approximation error is distributed uniformly across the rendered image.
While higher-order polynomials can effectively restore fidelity (with a third-order polynomial matching the quality of the original exponential), the increased computational complexity quickly reaches a point of diminishing returns. 
As polynomial orders increase, the arithmetic overhead can eventually outweigh the optimization benefit.

Our analysis indicates that the majority of the observed speed-up is attributed to tighter culling rather than function approximation alone. 
While frameworks such as gsplat and Faster-GS leverage native hardware instructions for fast exponential approximations, our polynomial approach remains a viable alternative for hardware lacking such specialized support. 
Importantly though, simply using a tighter culling as we do without changing the kernel leads to clearly visible artifacts as shown in Figure~\ref{fig:expartifacts}.

In our evaluation, we observed only one artifact that stands out and appears in Figure~\ref{fig:samples} most noticeably in the \emph{treehill} scene, manifesting as dark spots against the white sky.
This phenomenon is rooted in the unbounded nature of color values in vanilla Gaussian Splatting, which leads to color values far above the normalized white value of 1.
If the color would be clamped, the optimization would likely have increased the size of the splats or their opacity instead.
Our optimization assumes that peripheral areas with very low opacity can be safely culled without impacting visual quality.
However, when a splat's color intensity is sufficiently high, the product of its color and low peripheral opacity remains numerically significant.
Indeed, the rasterization hardware-accelerated approaches such as vk\_gaussian\_splatting exhibit similar artifacts, as the hardware blending clamps the color prior to blending as shown in Figure~\ref{fig:vktreehill}.
Notably, this issues is largely confined to high-intensity, overexposed background regions.

\subsection{Future Work}

While we discussed NPU specific details of our approach, an implementation on these architectures remains an objective for future research.
Therefore, further investigation efforts towards different NPU-like hardware, such as Tensor Cores, the Neural Engine or Huawei Ascend is necessary.
Additionally, relaxing the assumption that the polynomial kernels should fit the exponential allows for further investigation when optimizing with new kernels.
For example, the monotonicity issue associated with quadratic fits could be mitigated by enforcing a negative second-order coefficient $c_2$.
This constrained quadratic formulation is advantageous for NPU architectures like the Huawei Ascend 
as its 15-component vectors align almost perfectly with $16\times16$ matrix multiplication units, ensuring high hardware occupancy with minimal padding.
Finally, our approach is also applicable to the backward pass needed for optimization.
To which extent this influences training performance, memory consumption and densification, and how well it is compatible with the exponential kernel needs further investigation.

\section{Conclusion}
While the Gaussian kernel has long been the foundational 'de facto' choice for 3DGS-based rendering, 
we show that simplifying the evaluation by replacing the exponential function with a first-order polynomial inside a ReLU can lead to significant performance gains without any noticeable reduction in quality.

Our evaluation across standard datasets and multiple implementations of Gaussian Splatting demonstrates that approximating the traditional Gaussian kernel 
does not necessitate a complete overhaul of the rendering pipeline nor does it require re-optimizing existing Gaussian Splatting datasets.
On the contrary, it can be easily integrated into existing systems.
By focussing on the most computationally expensive part of the rendering process, reducing the amount of splats to be evaluated as well as simplifying the evaluation, 
we can achieve significant performance improvements of \perfimprovement{} while maintaining the same underlying mathematical framework for culling and anti-aliasing at virtually the same quality.
This suggests that the spatial contribution of a primitive is less dependent on the specific activation function used and more dependent on the underlying quadric support.

Ultimately, this work demonstrates that the efficiency of 3DGS can be significantly enhanced through pragmatic kernel approximation, 
enabling high-performance rendering on various platforms and hardware architectures without discarding the vast ecosystem of existing datasets.

\bibliographystyle{splncs04}
\bibliography{main}

\appendix

\section{Second-Order Approximation Range}

\begin{figure}[htbp]
	\centering
	\begin{subfigure}[b]{0.49\textwidth}
		\centering
		\includegraphics[width=\textwidth]{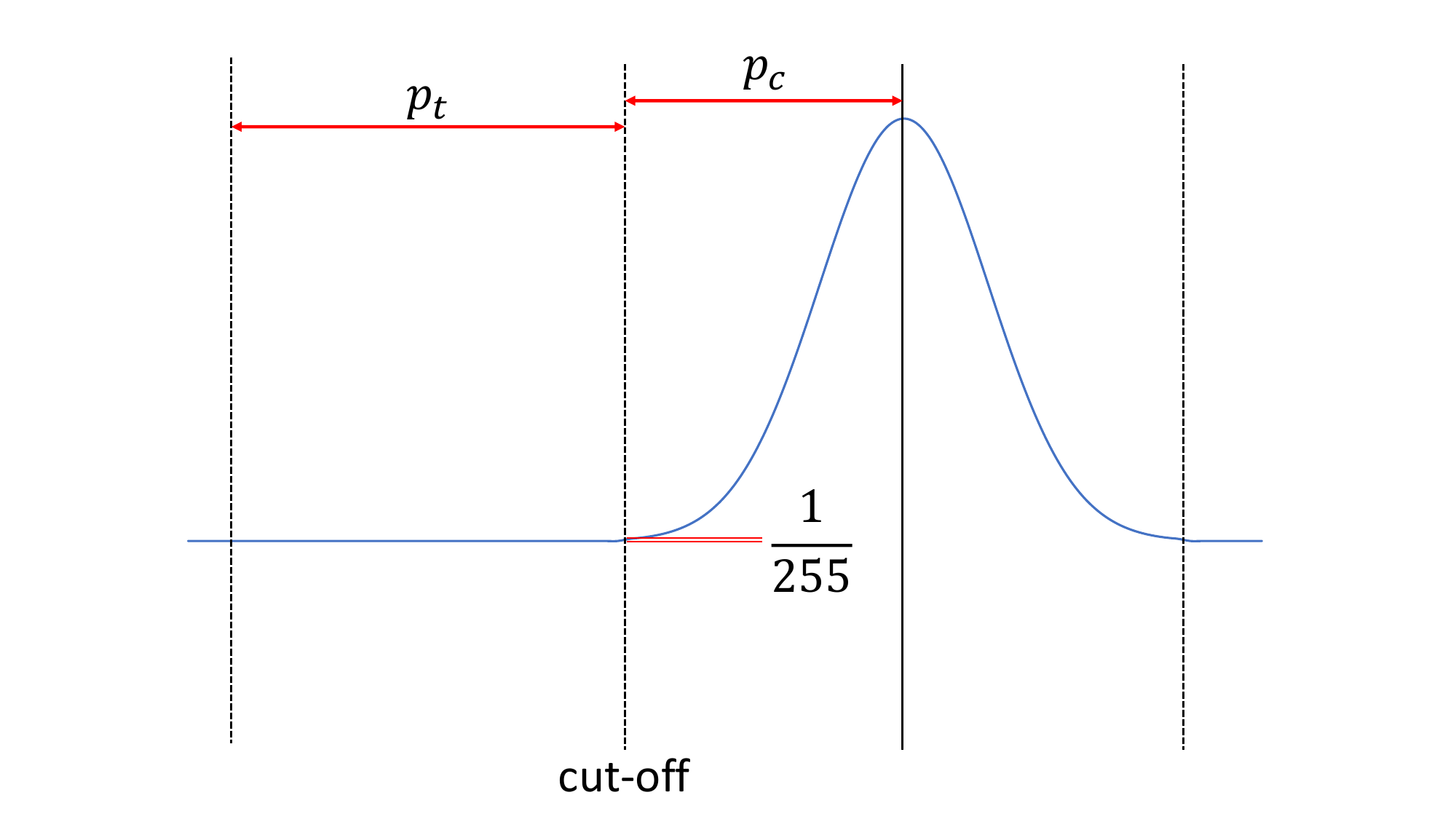}
		\caption{1D}
		\label{fig:Approx1D}
	\end{subfigure}
	\hfill
	\begin{subfigure}[b]{0.49\textwidth}
		\centering
		\includegraphics[width=\textwidth]{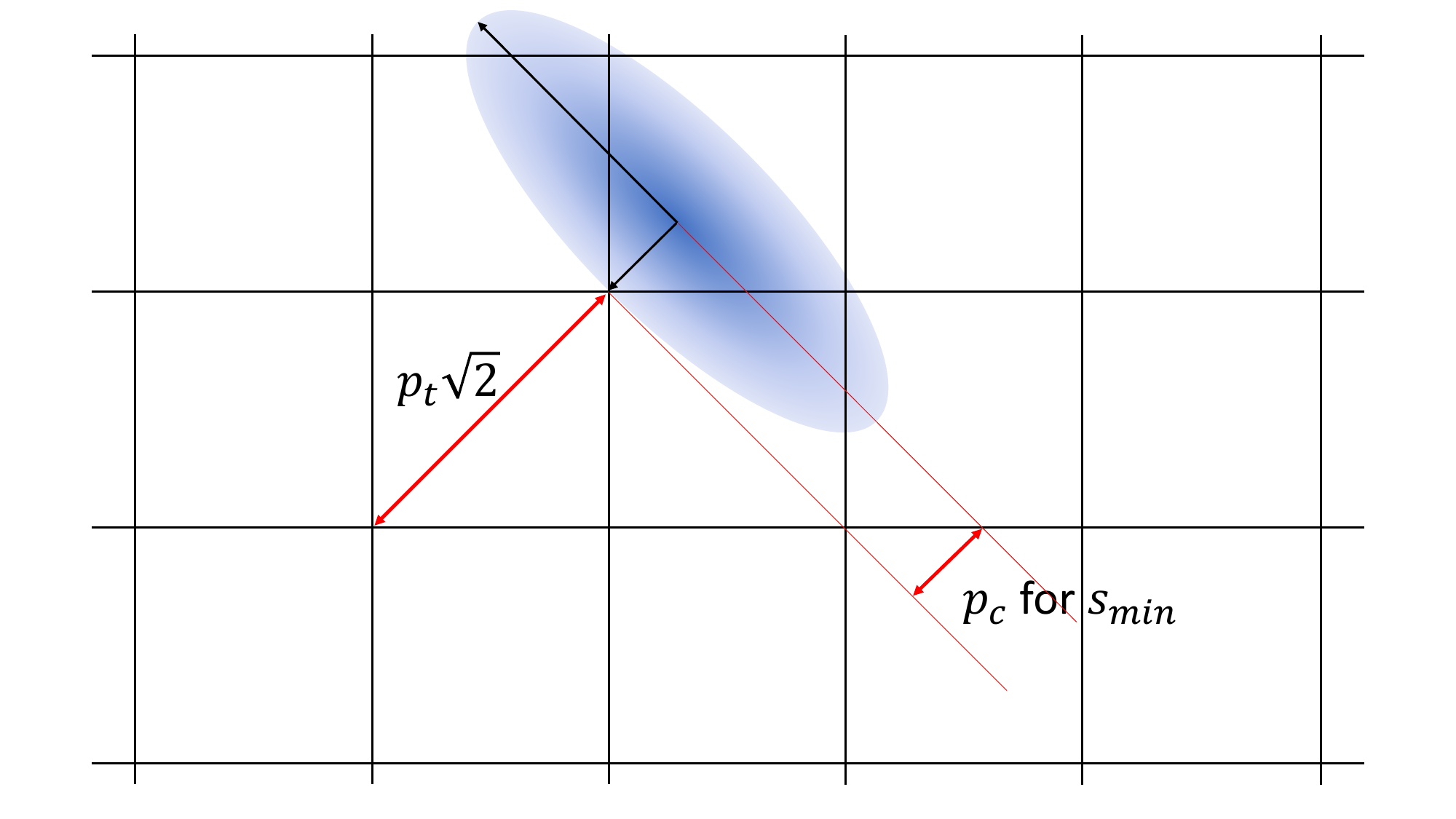}
		\caption{2D}
		\label{fig:Approx2D}
	\end{subfigure}
	\caption{To determine the range within which the kernel is evaluated we consider the worst case based on a tile based culling approach.
		The kernel is only evaluated for tiles which the splat touches.
		The worst case is when the splat just barely touches a tile in which case the kernel is still evaluated for the point within the tile furthest away from the splat center.
		In the one-dimensional case (a) this maximum distance to the cut-off $p_c$ plus the tile size $p_t$.
		In the two-dimensional case (b) $p_c$ is computed from the smaller splat scale and the diagonal of the tile needs to be added instead of just the tile size.
	}
	\label{fig:Approx}
\end{figure}

One solution for the monotonicity issue of the second-order polynomial is to increase the range in which we fit the polynomial to the exponential function to cover all possible values used for evaluation which depend on how splats are rasterized exactly.
Here, we look at a tile based approach where tiles are culled using the method presented in StopThePop\cite{radl2024stopthepop}.
This method culls tiles as tight as possible, \ie the splats are not evaluated on any tiles that the splat doesn't touch at all.
In the worst case this means that a splat barely touches a tile and the kernel is still evaluated for all pixels of the tile.

For explanatory purposes, we first look at the one-dimensional case, as depicted in Figure~\ref{fig:Approx1D}, where we define the Gaussian as $g_{\text{1D}}(p) = o\exp\left(-\frac{p^2}{2s^2}\right)$, where $s$ is the Gaussian's scale or standard deviation and $p$ is the distance of the pixel's center 
to the center of the splat $\mu$.
Considering the worst case scenario, we assume an opacity value of $o = 1$ and a splat that just barely touches the tile, leading to the need to evaluate the kernel for the whole tile.
The original Gaussian kernel is evaluated as non-zero from its center to the point where it is cut-off.
The distance $p_c$ from the splat's center to this cut-off position where $g_{\text{1D}}(p=p_c)$ reaches the value $\epsilon = \frac{1}{255}$, is computed as
\begin{equation}
	p_c = s \sqrt{-2 \ln\left(\epsilon\right)}.
\end{equation}%
Since the splat is evaluated for the whole tile even if it just barely touches one side, the point furthest away from the center is the point on the other side of the tile.
To factor this in, we have to add the tile size $p_t$ for our worst case scenario, leading to a maximum distance of $p_c + p_t$ for the one-dimensional case.

For the two-dimensional case, as depicted in Figure~\ref{fig:Approx2D}, the worst case we need to consider is a splat barely touching the corner of a tile and the maximum possible distance from that corner is given by the diagonal, which is $\sqrt{2}p_t$, in order to reach the opposite corner.
In two dimensions, we also need to consider the orientation of a given splat, as the worst case happens whenever the splat's smallest scale $s_{\min}$ is oriented 45 degrees along one of the diagonals while touching the corner of the tile.
For a fit that works for all splats, we need to find the minimum $s_{\min}$ of all splats in a scene.
With anti-aliasing, which adds a value $v$ to $s^2$, there is a general lower limit on the smallest possible $s_{\min} = \sqrt{v}$ for all splats.
With this worst case in mind, we can compute the upper limit $x_{\max}$ for the range within which the approximation of $g(x)$ has to be computed for $f_2(x)$ as
\begin{equation}
	x_{\max} = \left(\frac{p_c + \sqrt{2}p_t}{s_{\min}}\right)^2 = \left(\frac{\sqrt{2}p_t}{s_{\min}} + \sqrt{-2\ln\left(\epsilon\right)}\right)^2.
\end{equation}%
Because $x_{\max}$ scales quadratically with the tile size, larger tiles can cause the second-order fit to become practically indistinguishable from the first-order fit, as the range of $x$ values, for which the second-order polynomial must be accurate, becomes excessively large.
Consequently, maintaining a small tile size is critical for a practical fit of a second-order polynomial that has better quality than $f_1(x)$ with this approach.

\section{Higher-Order Culling}

The higher-order culling equations are derived from the standard formulas for quadratic equations for $f_2(x)$ to compute the threshold
\begin{equation}
	t_{f_2} = \frac{-c_1 - \sqrt{c_1^2 - 4\left(c_0 - \frac{\epsilon}{o}\right)c_2}}{2c_2},
\end{equation}%
and the general solution for cubic equations for the threshold of $f_3(x)$
\begin{eqnarray}
	\Delta_0 &=& c_2^2 - 3 c_3c_1 \\
	\Delta_1 &=& 2c_2^3 - 9c_3c_2c_1+27c_3^2\left(c_0-\frac{\epsilon}{o}\right) \\
	C &=& \sqrt[3]{\frac{\Delta_1 + \sqrt{\Delta_1^2 - 4\Delta_0^3}}{2}} \\
	t_{f_3} &=& -\frac{1}{3c_3}\left(c_2 + C + \frac{\Delta_0}{C}\right),
\end{eqnarray}%
taking the first root that is real and bigger than 0 for each polynomial.
Just as for $f_1(x)$, we can compute a general threshold for all splats by setting $\epsilon=0$ and thus computing an opacity-independent threshold valid for all splats.
Otherwise the term $c_0 - \frac{\epsilon}{o}$ appears as opacity-dependent term and can be used for tighter, opacity-aware culling where each splat has its own threshold.

\section{Detailed Timings for Higher-Order Polynomials}

Table~\ref{tab:timing_supplementaly} shows performance of Baseline and Faster-GS for the higher-order polynomials $f_2(x)$ and $f_3(x)$.
As expected, the performance gains are lower in comparison to $f_1(x)$, but potentially still practical.
Across both approaches, $f_3(x)$ was slower than $g(x)$ by only $1\%$ in a single scene.
Consequently, extending the model beyond the third order is unnecessary, as it would likely decrease performance further without offering quality improvements over $f_3(x)$, which already matches $g(x)$.
\begin{table}
	\centering
	\caption{Performance measurements of our first-, second- and third-order approximation in Baseline and Faster-GS.
	Timing measurements in the kernel columns are the frame time in milliseconds, measured on the GPU.
	The \% columns show the performance improvement of the kernel left of it over $g(x)$ in percent.}
	\label{tab:timing_supplementaly}
	\begin{tabular}{l|rrrrrrr|rrrrrrr}
\toprule
 & \multicolumn{7}{c}{Baseline} & \multicolumn{7}{c}{Faster-GS} \\
 & $g(x)$ & $f_1(x)$ & \hspace{0.5em}\% & $f'_2(x)$ & \hspace{0.5em}\% & $f_3(x)$ & \hspace{0.5em}\% & $g(x)$ & $f_1(x)$ & \hspace{0.5em}\% & $f'_2(x)$ & \hspace{0.5em}\% & $f_3(x)$ & \hspace{0.5em}\% \\
\midrule
bicycle & 3.74 & 3.29 & 12 & 3.62 & 3 & 3.77 & -1 & 2.55 & 2.38 & 7 & 2.44 & 4 & 2.52 & 1 \\
bonsai & 2.40 & 2.07 & 14 & 2.21 & 8 & 2.32 & 3 & 0.85 & 0.75 & 11 & 0.81 & 5 & 0.85 & 0 \\
counter & 2.24 & 1.59 & 29 & 1.74 & 22 & 2.07 & 8 & 1.01 & 0.87 & 14 & 0.94 & 7 & 1.00 & 1 \\
drjohnson & 2.43 & 2.03 & 17 & 2.14 & 12 & 2.37 & 2 & 1.20 & 1.10 & 8 & 1.16 & 3 & 1.19 & 1 \\
flowers & 3.00 & 2.38 & 21 & 2.60 & 13 & 2.85 & 5 & 1.57 & 1.43 & 9 & 1.49 & 5 & 1.55 & 1 \\
garden & 2.67 & 2.15 & 20 & 2.22 & 17 & 2.55 & 5 & 2.37 & 2.21 & 7 & 2.25 & 5 & 2.33 & 2 \\
kitchen & 2.39 & 1.96 & 18 & 2.17 & 9 & 2.32 & 3 & 1.64 & 1.46 & 11 & 1.56 & 4 & 1.64 & 0 \\
playroom & 1.76 & 1.42 & 19 & 1.50 & 14 & 1.66 & 6 & 0.87 & 0.76 & 13 & 0.81 & 7 & 0.85 & 2 \\
room & 2.61 & 1.93 & 26 & 2.09 & 20 & 2.42 & 7 & 0.92 & 0.78 & 15 & 0.84 & 8 & 0.91 & 1 \\
stump & 2.95 & 2.54 & 14 & 2.67 & 9 & 2.86 & 3 & 1.43 & 1.32 & 8 & 1.40 & 2 & 1.45 & -1 \\
train & 2.06 & 1.78 & 14 & 1.85 & 10 & 1.97 & 5 & 1.21 & 1.12 & 7 & 1.16 & 4 & 1.19 & 1 \\
treehill & 2.62 & 2.21 & 15 & 2.35 & 10 & 2.59 & 1 & 1.52 & 1.36 & 10 & 1.43 & 6 & 1.51 & 1 \\
truck & 1.87 & 1.64 & 12 & 1.70 & 9 & 1.83 & 2 & 1.42 & 1.36 & 4 & 1.39 & 2 & 1.41 & 1 \\
average & 2.52 & 2.08 & 18 & 2.22 & 12 & 2.43 & 4 & 1.43 & 1.30 & 10 & 1.36 & 5 & 1.41 & 1 \\
\bottomrule
\end{tabular}

\end{table}

\section{Detailed Optimization Results}
Our ablation demonstrates that optimization using polynomial kernels improves both quality and performance compared to using the polynomial kernels as a drop-in replacement for inference only.
Table~\ref{tab:optimization} provides a per-scene breakdown of these metrics, including the total splat count for each scene.
Although the number of splats generally increases with optimization, this does not result in a performance penalty.
In fact, the ablation reveals a net performance gain, implying that the average spatial coverage of the splats must decrease to compensate for the higher count.

\begin{sidewaystable}
	\centering
	\caption{Quality comparison of the different kernels using the quality metrics PSNR, SSIM and LPIPS.
	For LPIPS lower is better, for the other two higher is better.}
	\label{tab:optimization}
	\scriptsize
	\begin{tabular}{l|rrrr|rrrr|rrrr|rrrr|rrrr}
\toprule
 & \multicolumn{4}{c}{PSNR $\uparrow$} & \multicolumn{4}{c}{SSIM $\uparrow$} & \multicolumn{4}{c}{LPIPS $\downarrow$} & \multicolumn{4}{c}{Baseline} & \multicolumn{4}{c}{Splats} \\
 & $g(x)$ & $f_1(x)$ & $f'_2(x)$ & $f_3(x)$ & $g(x)$ & $f_1(x)$ & $f'_2(x)$ & $f_3(x)$ & $g(x)$ & $f_1(x)$ & $f'_2(x)$ & $f_3(x)$ & $g(x)$ & $f_1(x)$ & $f'_2(x)$ & $f_3(x)$ & $g(x)$ & $f_1(x)$ & $f'_2(x)$ & $f_3(x)$ \\
\midrule
bicycle & 25.06 & 24.93 & 25.00 & 25.19 & 0.749 & 0.730 & 0.737 & 0.748 & 0.243 & 0.272 & 0.263 & 0.245 & 3.737 & 3.563 & 3.506 & 3.753 & 4902288 & 5343544 & 5171329 & 4946096 \\
bonsai & 32.42 & 32.34 & 32.32 & 32.42 & 0.946 & 0.944 & 0.945 & 0.947 & 0.180 & 0.180 & 0.180 & 0.179 & 2.397 & 1.758 & 1.739 & 2.213 & 1067589 & 1118872 & 1114658 & 1068086 \\
counter & 29.10 & 29.13 & 29.13 & 29.16 & 0.916 & 0.912 & 0.914 & 0.915 & 0.182 & 0.186 & 0.185 & 0.182 & 2.242 & 1.979 & 2.098 & 2.054 & 1045285 & 1081086 & 1078830 & 1057410 \\
drjohnson & 29.43 & 29.43 & 29.43 & 29.46 & 0.906 & 0.905 & 0.906 & 0.906 & 0.235 & 0.239 & 0.237 & 0.235 & 2.426 & 2.341 & 2.292 & 2.331 & 3172750 & 3150256 & 3101417 & 3170250 \\
flowers & 21.37 & 21.24 & 21.26 & 21.40 & 0.587 & 0.577 & 0.579 & 0.587 & 0.362 & 0.366 & 0.366 & 0.360 & 3.001 & 2.613 & 2.803 & 2.599 & 2982659 & 3431313 & 3303908 & 3051499 \\
garden & 27.29 & 26.96 & 27.09 & 27.28 & 0.854 & 0.841 & 0.845 & 0.853 & 0.130 & 0.150 & 0.147 & 0.132 & 2.671 & 2.622 & 2.577 & 2.628 & 4585948 & 5024111 & 5037249 & 4702603 \\
kitchen & 31.62 & 31.21 & 31.38 & 31.37 & 0.933 & 0.929 & 0.931 & 0.932 & 0.115 & 0.118 & 0.117 & 0.116 & 2.390 & 2.307 & 2.475 & 2.365 & 1536135 & 1638896 & 1628337 & 1581734 \\
playroom & 30.21 & 29.97 & 30.20 & 30.28 & 0.910 & 0.906 & 0.909 & 0.911 & 0.239 & 0.244 & 0.243 & 0.239 & 1.756 & 1.849 & 1.755 & 1.694 & 1838144 & 1924734 & 1885679 & 1861425 \\
room & 31.77 & 31.41 & 31.63 & 31.81 & 0.928 & 0.924 & 0.926 & 0.928 & 0.196 & 0.202 & 0.200 & 0.196 & 2.609 & 2.214 & 2.417 & 2.285 & 1259659 & 1290728 & 1275659 & 1267053 \\
stump & 26.56 & 26.39 & 26.33 & 26.57 & 0.767 & 0.757 & 0.758 & 0.767 & 0.244 & 0.258 & 0.255 & 0.243 & 2.951 & 2.497 & 2.445 & 2.656 & 4100374 & 4452758 & 4412930 & 4172081 \\
train & 22.08 & 21.98 & 21.79 & 22.00 & 0.819 & 0.811 & 0.814 & 0.819 & 0.202 & 0.211 & 0.208 & 0.202 & 2.065 & 1.960 & 1.937 & 1.999 & 1089477 & 1252397 & 1182744 & 1108791 \\
treehill & 22.69 & 22.43 & 22.60 & 22.55 & 0.637 & 0.627 & 0.631 & 0.637 & 0.349 & 0.368 & 0.362 & 0.351 & 2.616 & 2.677 & 2.601 & 2.776 & 3065667 & 3413884 & 3212054 & 3064993 \\
truck & 25.45 & 25.20 & 25.26 & 25.42 & 0.884 & 0.881 & 0.882 & 0.884 & 0.148 & 0.153 & 0.152 & 0.148 & 1.866 & 1.889 & 1.866 & 1.904 & 2955345 & 3567014 & 3373271 & 3033360 \\
\bottomrule
\end{tabular}

\end{sidewaystable}

\ifSubfilesClassLoaded{%
    \newpage
    \bibliography{main}
    \bibliographystyle{splncs04}
}{}
\end{document}

\end{document}
\typeout{get arXiv to do 4 passes: Label(s) may have changed. Rerun}